\documentclass[journal]{IEEEtran}

\ifCLASSINFOpdf
\else
   \usepackage[dvips]{graphicx}
\fi
\usepackage{cite}
\usepackage{amsmath,amssymb,amsfonts}
\usepackage{algorithmic}
\usepackage{graphicx}
\usepackage{textcomp}
\usepackage{xcolor}
\usepackage[hidelinks]{hyperref}

\ifCLASSINFOpdf
\else
\fi

\usepackage{amssymb}
\usepackage{multirow}

\hypersetup{hypertex=true,
            colorlinks=true,
            linkcolor=cyan,
            citecolor=cyan,
            urlcolor=black}
\usepackage{url}

\usepackage{graphicx}
\usepackage{booktabs}
\usepackage{color}
\usepackage[justification=centering,font=footnotesize,labelsep=period]{caption}
\begin{document}
%

% paper title
% Titles are generally capitalized except for words such as a, an, and, as,
% at, but, by, for, in, nor, of, on, or, the, to and up, which are usually
% not capitalized unless they are the first or last word of the title.
% Linebreaks \\ can be used within to get better formatting as desired.
% Do not put math or special symbols in the title.
\title{Zero-Shot Low-Light Image Enhancement via Joint Frequency Domain Priors Guided Diffusion}
% Joint Wavelet and Fourier Priors Guided Diffusion for Zero-Shot Low-Light Image Enhancement
% Joint frequency domain
% author names and affiliations
% transmag papers use the long conference author name format.

\author{\IEEEauthorblockN{Jinhong He, Shivakumara Palaiahnakote, Aoxiang Ning, Minglong Xue\IEEEauthorrefmark{1}}

%\author{\IEEEauthorblockN{Minglong Xue\IEEEauthorrefmark{1},
%Yanyi He\IEEEauthorrefmark{2},
%Senming Zhong\IEEEauthorrefmark{3}, 
%Montgomery Scott\IEEEauthorrefmark{3}, and
%Eldon Tyrell\IEEEauthorrefmark{4},~\IEEEmembership{Fellow,~IEEE}}
% \IEEEauthorblockA{\IEEEauthorrefmark{1}School of Electrical and Computer Engineering,
% Georgia Institute of Technology, Atlanta, GA 30332 USA}
% \IEEEauthorblockA{\IEEEauthorrefmark{2}Twentieth Century Fox, Springfield, USA}
% \IEEEauthorblockA{\IEEEauthorrefmark{3}Starfleet Academy, San Francisco, CA 96678 USA}
% \IEEEauthorblockA{\IEEEauthorrefmark{4}Tyrell Inc., 123 Replicant Street, Los Angeles, CA 90210 USA}% <-this % stops an unwanted space
\thanks{
This work is supported by the Science and Technology Research Program of Chongqing Municipal Education Commission (KJQN202401106), the Special Project for the Central Government to Guide Local Science and Technology Development (2024ZYD0334), Chongqing Postgraduate Research and Innovation Project Funding (Grant No. CYS240680).\par (Corresponding author: Minglong Xue)
Jinhong He, Aoxiang Ning and Minglong Xue are with Chongqing University of Technology, Chongqing, 400054, China. (e-mail:  hejh@stu.cqut.edu.cn, xueml@cqut.edu.cn, ningax@stu.cqut.edu.cn) Shivakumara Palaiahnakote is with the School of Science, Engineering \& Environment University of Salford, Manchester, UK. (e-mail: S.Palaiahnakote@salford.ac.uk)}
}
% \thanks{Manuscript received xx, xx; revised August 26, 2015. 
% Corresponding author: M. Shell (email: http://www.michaelshell.org/contact.html).}

% The paper heade
% \markboth{Journal of \LaTeX\ Class Files,~Vol.~14, No.~8, August~2015}%
% {Shell \MakeLowercase{\textit{et al.}}: Bare Demo of IEEEtran.cls for IEEE Transactions on Magnetics Journals}

% The only time the second header will appear is for the odd numbered pages
% after the title page when using the twoside option.
% 
% *** Note that you probably will NOT want to include the author's ***
% *** name in the headers of peer review papers.                   ***
% You can use \ifCLASSOPTIONpeerreview for conditional compilation here if
% you desire.

% If you want to put a publisher's ID mark on the page you can do it like
% this:
%\IEEEpubid{0000--0000/00\$00.00~\copyright~2015 IEEE}
% Remember, if you use this you must call \IEEEpubidadjcol in the second
% column for its text to clear the IEEEpubid mark.

% use for special paper notices
%\IEEEspecialpapernotice{(Invited Paper)}

% for Transactions on Magnetics papers, we must declare the abstract and
% index terms PRIOR to the title within the \IEEEtitleabstractindextext
% IEEEtran command as these need to go into the title area created by
\maketitle
% As a general rule, do not put math, special symbols or citations
% in the abstract or keywords.
% \IEEEtitleabstractindextext{%

\begin{abstract}
Due to the singularity of real-world paired datasets and the complexity of low-light environments, this leads to supervised methods lacking a degree of scene generalisation. Meanwhile, limited by poor lighting and content guidance, existing zero-shot methods cannot handle unknown severe degradation well. To address this problem, we will propose a new zero-shot low-light enhancement method to compensate for the lack of light and structural information in the diffusion sampling process by effectively combining the wavelet and Fourier frequency domains to construct rich a priori information. The key to the inspiration comes from the similarity between the wavelet and Fourier frequency domains: both light and structure information are closely related to specific frequency domain regions, respectively. Therefore, by transferring the diffusion process to the wavelet low-frequency domain and combining the wavelet and Fourier frequency domains by continuously decomposing them in the inverse process, the constructed rich illumination prior is utilised to guide the image generation enhancement process. Sufficient experiments show that the framework is robust and effective in various scenarios. The code will be available at: \href{https://github.com/hejh8/Joint-Wavelet-and-Fourier-priors-guided-diffusion}{https://github.com/hejh8/Joint-Wavelet-and-Fourier-priors-guided-diffusion}.
\end{abstract}

% Note that keywords are not normally used for peerreview papers.

% make the title area
% \maketitle

% To allow for easy dual compilation without having to reenter the
% abstract/keywords data, the \IEEEtitleabstractindextext text will
% not be used in maketitle, but will appear (i.e., to be "transported")
% here as \IEEEdisplaynontitleabstractindextext when the compsoc 
% or transmag modes are not selected <OR> if conference mode is selected 
% - because all conference papers position the abstract like regular
% papers do.
\IEEEdisplaynontitleabstractindextext
% \IEEEdisplaynontitleabstractindextext has no effect when using
% compsoc or transmag under a non-conference mode.

% For peer review papers, you can put extra information on the cover
% page as needed:
% \ifCLASSOPTIONpeerreview
% \begin{center} \bfseries EDICS Category: 3-BBND \end{center}
% \fi
%
% For peerreview papers, this IEEEtran command inserts a page break and
% creates the second title. It will be ignored for other modes.
\IEEEpeerreviewmaketitle

\section{Introduction}
Damage to the quality of captured images due to real-world low-light conditions will affect the development of various computer vision applications. This has attracted the attention of researchers and a series of studies have been conducted to stabilise the performance of various downstream vision tasks \cite{li2021deep,xue2020arbitrarily}. Low-light image enhancement aims to reverse various degraded domains and restore the original clean image by learning the mapping relationship between degraded and normal domains. In the past decades, traditional methods have contributed significant results to the field of low-light image enhancement by optimising the parameters of the image itself \cite{guo2016lime}. However, these hand-crafted a priori lack sufficient adaptivity, resulting in enhancement results with serious instability and performance gaps. 

With the development of science, research on low-light image enhancement based on deep learning has made significant progress. However, current research \cite{xue2024dldiff,ning2024kan} still focuses on fitting real-world lighting conditions using a large amount of paired data. Due to the difficulty in obtaining real-world paired data and model-specific debugging methods, this leads to models that are highly dependent on supervision of paired data and lack the ability to generalise to the real world. 

% \begin{figure}[t!]\centering
%      \includegraphics[height=0.35\textwidth,width=0.48\textwidth]{duibi0.pdf}	    
%      \caption{Our method effectively learns the nonlinear degradation factors in the low-light domain, especially in darker scenes, and our recovery significantly improves compared to the GSAD.}
%     \label{sample}
% \vspace{-5mm}
% \end{figure}

\begin{figure}[t!]
\centering
    \includegraphics[height=0.3\textwidth,width=0.42\textwidth]{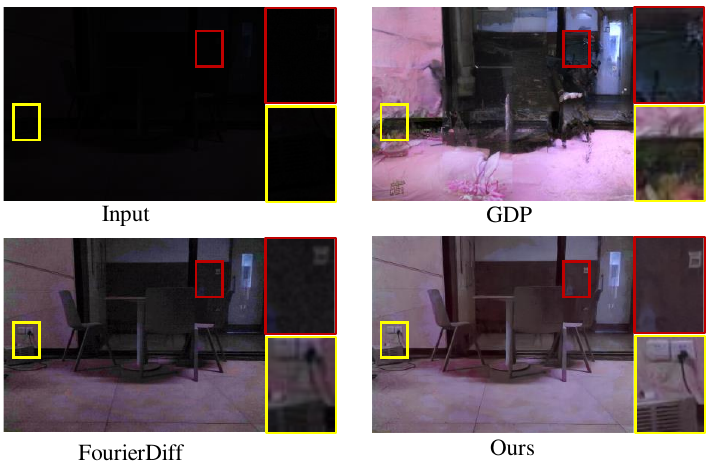}	    
\caption{Visual comparison with SOTA diffusion-based zero-shot methods.}
\label{1}
\vspace{-5mm}
\end{figure}

Based on the above problems, it has triggered researchers to focus on unsupervised enhancement methods with remarkable results \cite{guo2020zero,fu2023learning,jiang2021enlightengan,yang2023implicit}. In this process, the generative models based on promoting better perceptual quality for low-light image enhancement gained some recognition, the recently emerged diffusion model \cite{ho2020denoising} became one of the hotspots because of its remarkable generative effect and demonstrated its effectiveness in the field of image enhancement through supervised methods \cite{xue2024dldiff,xue2024low}. To connect unsupervised learning and diffusion models to improve their generality in the real world, \cite{jiang2024lightendiffusion,he2024zero} construct content-rich reflectance maps based on retinex theory to facilitate unsupervised recovery. Further, some methods \cite{fei2023generative,lv2024fourier} employ zero-shot approaches to enhance model generalisation capabilities by pre-training diffusion models with good prior. However, as shown in Fig. \ref{1}, these methods are limited by the paucity of a priori information about lighting and content, and thus tend to perform poorly in complex and unknown real-world scenarios. 

To further mitigate the above challenges, inspired by \cite{lv2024fourier}, we perform a preliminary validation. The key to the inspired validation lies in the similarity between the wavelet domain and the Fourier frequency domain: both the wavelet low-frequency domain and the Fourier amplitude concentrate the lighting information of the image, while the wavelet high-frequency domain and the Fourier phase concentrate the structural information of the image. Specifically, the wavelet-decomposed low-frequency domain has a better exposure effect \cite{jiang2023low} compared to the degraded image, and in combination with the Fourier amplitude of the normal image, it can further direct the lighting information of the image. In addition, previous work \cite{si2024freeu} demonstrated that high-frequency information is more susceptible to damage during diffusion, and by retaining the high-frequency features the image structure can be further preserved, facilitating the generation of consistent data distributions from the enhancement results.
% As shown in Fig. (#), it can be easily seen that the low-frequency domain after wavelet decomposition has a better exposure compared to the degraded image, and the light information of the image can be further guided by combining the Fourier magnitude of the normal image. In addition, previous work \cite{si2024freeu} demonstrated that high-frequency information is more susceptible to damage during diffusion, and by retaining the high-frequency features the image structure can be further preserved, facilitating the generation of consistent data distributions from the enhancement results.

In this letter, we propose a new zero-shot low-light image enhancement method to balance the degradation of light and structure by effectively connecting the wavelet domain and the Fourier frequency domain to construct rich a priori information embedded in the pre-trained diffusion model to compensate for the lack of light and structural information in the diffusion generation process in zero-shot enhancement. Specifically, we first transfer the diffusion process to the wavelet low-frequency domain of the degraded image, preserving the high-frequency information. Subsequently, the wavelet domain and Fourier frequency domain are continuously decomposed in the inverse process of multiple steps, and by updating the combination of wavelet information and Fourier amplitude of the enhancement result and wavelet information and Fourier phase of the degraded image, we gradually make up for the impoverished illumination and structural information in the a priori information, and promote the alignment with the data distribution of the natural image. Meanwhile, to strengthen the constraints on the generation process, we introduce a multimodal text embedding inverse process that iteratively steers the sampling results closer to positively orientated enhancements. Notably, we simplify a generic denoising method \cite{li2023fastllve,panagiotou2024denoising} to address the legacy of noise for refinement, which further improves performance. Sufficient experiments show that the framework is generalisable in a variety of scenarios. Our main contributions are summarised below:
\begin{itemize}
% \item We propose a new zero-shot low-light image enhancement method, which effectively compensates for the paucity of light and structural information in the zero-shot enhancement process by combining the wavelet and Fourier frequency domains to construct a rich a priori information embedded in a pre-trained diffusion model.
\item We propose a zero-shot low-light image enhancement method, which effectively compensates for the lack of light and structural information in the zero-shot diffusion enhancement process by constructing rich prior information in the joint wavelet and Fourier frequency domains.
\item We further explore the effective combination of wavelet and Fourier frequency domains to satisfy the degradation and enhancement for different complex scenes. Meanwhile, we also introduce multimodal text embedding to guide the positive enhancement direction.
\item Extensive experiments have demonstrated the effectiveness and versatility of the proposed method.
% In extensive experiments based on real-world datasets, we demonstrate the effectiveness and generalisation of the proposed method. 
\end{itemize}

\begin{figure*}[ht!]\centering
    \includegraphics[height=0.45\textwidth,width=1\textwidth]{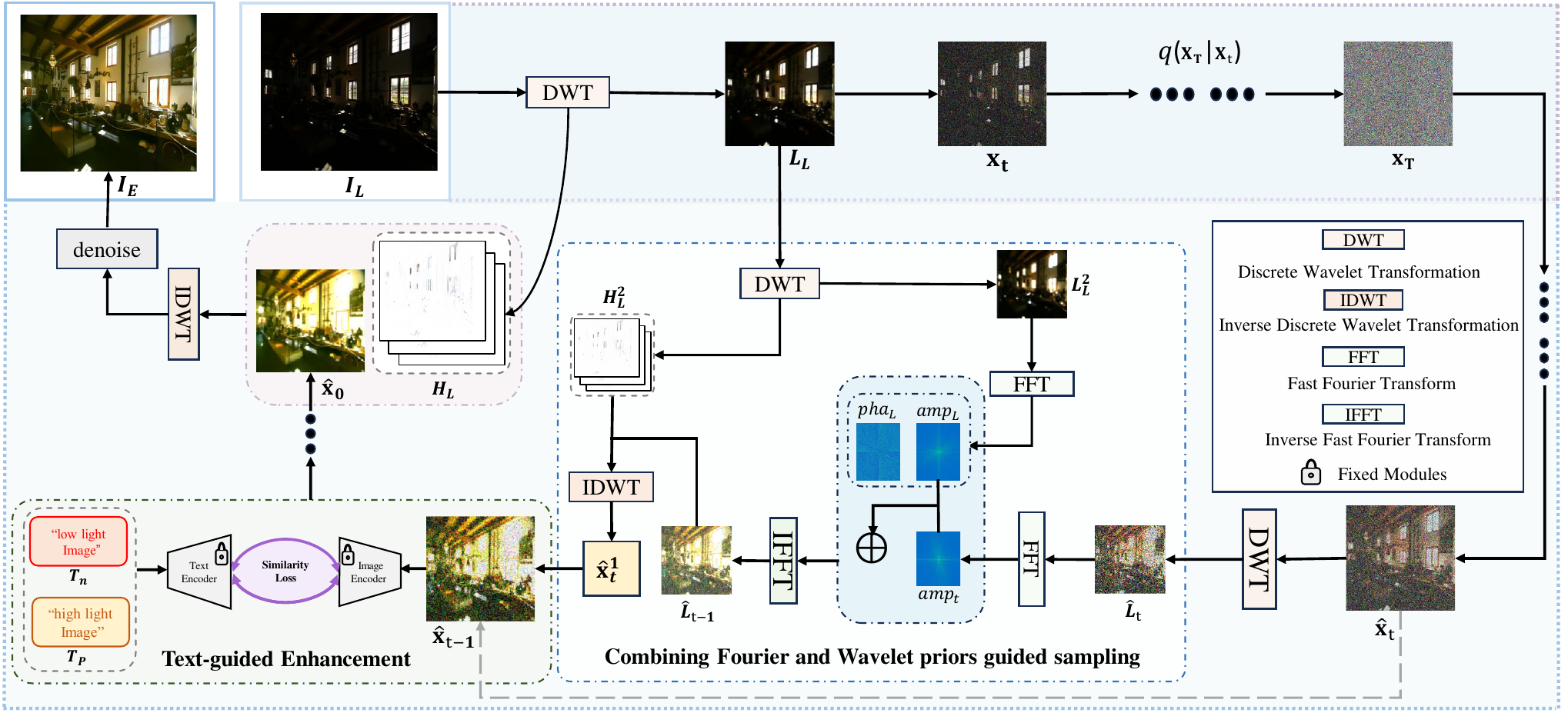}	    
\caption{Detailed structure and flow of the proposed method.}
\label{frame}
\vspace{-5mm}
\end{figure*}
% Learn the parameters of image degradation due to noise and use the learned parameters to generate the image. 

\section{Methods}
\subsection{Diffusion Model Preliminary}
The diffusion model corresponds to a Markov chain structure. It consists of two main steps: the forward diffusion process $q$ and the backsampling process $p_\theta$. The diffusion model forward process works by gradually adding Gaussian noise to the real input image $x_0$ until the time steps of $T$ approximate purely noisy data $x_T$. Since the noise added to the forward process is independent and follows a normal distribution with mean $\sqrt{1-\beta_t}x_{(t-1)}$ and variance $\beta_tI$. This process can be simplified as:

\begin{equation}
q(x_t|x_0)=N(x_t;\sqrt{\overline{\alpha}_t}x_0,({1-\overline{\alpha}_t})I),
\tag{1}
\end{equation}
where $\alpha_t=1-\beta_t$, $\overline{\alpha}_t$=$\prod_{i=1}^{t}\alpha_i$, $t\in\{[1,...T\}$. In the inverse sampling process, we recover the image $\hat{x}_0$ by gradually denoising the pure noise $x_T$. The inverse sampling process is denoted as:
\begin{equation}\label{3}
p_\theta(\hat{x}_{t-1}|\hat{x}_t)=N(\hat{x}_{t-1};\mu _\theta(\hat{x}_t,t),\sigma ^2_tI),
\tag{2}
\end{equation}
where $\mu _\theta=\frac{1}{\sqrt{\alpha_t}}(x_t-\frac{\beta_t}{\sqrt{1-\overline{\alpha}_t}}\epsilon_\theta(x_t,t))$ is the diffusion model noise predictor, $\epsilon_\theta(x_t,t)$ is the estimated noise
% \begin{equation}
% \mu _\theta=\frac{1}{\sqrt{\alpha_t}}(x_t-\frac{\beta_t}{\sqrt{1-\overline{\alpha}_t}}\epsilon_\theta(x_t,t)),
% \tag{3}
% \end{equation}
% where $\epsilon_\theta(x_t,t)$ is the estimated noise.

\subsection{Joint Wavelet and Fourier priors guided sampling}
Due to the highly stochastic nature of the diffusion model in the sampling process, this leads to extremely high demands on the constraints of the sampling process. Meanwhile, previous work \cite{si2024freeu} demonstrated the sensitivity of the diffusion model to high-frequency information. Therefore, how to enhance the guidance of light information while maintaining the data distribution remains an important challenge for zero-shot low-light enhancement. Inspired by \cite{lv2024fourier}, we extract the generative prior via a pre-trained diffusion model and combine wavelet and Fourier frequency domains to construct a rich light prior to guide the sampling process to produce visually friendly and data-consistent results.

As shown in Fig. \ref{frame}, we first perform discrete wavelet transform $DWT$ on the input image $I_L$ to obtain its low-frequency domain $L_L$ and high-frequency domain $H_L$, and transfer the diffusion process to the wavelet low-frequency domain $L_L$. Meanwhile, for $L_L$, we again perform the frequency-domain decomposition via wavelet transform to retain its low-frequency $L_L^2$ and high-frequency information $H_L^2$, which are used to construct the a priori and guide the updating of the sampling process. It can be expressed as:
\begin{equation}
\{L_L,H_L\}={ DWT}(I_L),
\tag{4}
\end{equation}
\begin{equation}
\{L_L^2,H_L^2\}={ DWT}(L_L).
\tag{5}
\end{equation}

Subsequently, we perform wavelet frequency-domain decomposition and Fourier amplitude-phase decomposition for each sampling result $\hat{x}_{t}$ of the inverse sampling process. Specifically, we first perform a wavelet transform on each sampling result $\hat{x}_{t}$ to obtain its low-frequency domain $\hat{L}_{t}$, and then perform a Fourier transform ($FFT$) on its $\hat{L}_{t}$ and $L_L^2$, respectively, to obtain the amplitude and phase, as follows:
\begin{equation}
\{\hat{L}_{t},\hat{H}_{t}\}={DWT}(\hat{x}_{t}),
\tag{6}
\end{equation}
\begin{equation}
amp_{t},pha_{t}=FFT(\hat{L}_{t}),
\tag{7}
\end{equation}
\begin{equation}
amp_{L},pha_{L}=FFT(L_L^2),
\tag{8}
\end{equation}
where $amp_{t}$, $amp_{L}$, $pha_{t}$, and $pha_{L}$ denote the amplitude and phase of $\hat{L}_{t}$ and $L_L^2$, respectively. As mentioned earlier, the wavelet low-frequency domain contains the luminance information and most of the structural information of the image, while the Fourier amplitude contains the luminance information of the image. We combine the wavelet low-frequency domain with the Fourier amplitude and update the amplitude using $amp_{t}$ and $amp_{L}$ to construct a rich luminance prior. In addition, the wavelet high-frequency information and phases extracted from the sampling results are more likely to produce random details that affect the data distribution of the image. As shown in previous work \cite{si2024freeu}, high-frequency information is more susceptible to interference during the diffusion process. Therefore, we replace $\hat{H}_{t}$ and $pha_{t}$ with the high-frequency details $H_L^2$ and the phase $pha_{L}$ of the input image $I_L$ to guide the content generation of the sampling process and ensure the fidelity and consistency of the data distribution. Next, we obtain the updated sampling result $\hat{x}_t^1$ using the Inverse Fast Fourier Transform ($IFFT$) and the Inverse Discrete Wavelet Transform ($IDWT$), which is defined as:
\begin{equation}
\hat{x}_t^1=IDWT(IFFT(\vartheta amp_{t}+amp_{L},pha_{L}),H_L^2),
\tag{9}
\end{equation}
where $\vartheta$ denotes the luminance level learnable factor that controls the sampling process. At this point, we follow the setup of \cite{wang2022zero} to obtain the next sampling result from the joint distribution, denoted as:
\begin{equation}\label{3}
p_\theta(\hat{x}_{t-1}|\hat{x}_t,\hat{x}_t^1)=N(\hat{x}_{t-1};\mu_t(\hat{x}_t,\hat{x}_t^1),\sigma ^2_tI),
\tag{10}
\end{equation}
where $\mu_t(\hat{x}_t,\hat{x}_t^1)=\frac{\sqrt{\overline{\alpha}_{t-1}}\beta_t}{1-\overline{\alpha}_t}\hat{x}_t^1 +\frac{\sqrt{\alpha_t}(1-\overline{\alpha}_{t-1})}{1-\overline{\alpha}_t}\hat{x}_t$.

By bootstrapping the sampling process iteratively using a joint prior constructed from wavelets and Fourier, we finally obtain the result $\hat{x}_{0}$, which has a good luminance distribution. To restore the image and further maintain the consistency of the data distribution, we perform an inverse wavelet transform on the sampling result $\hat{x}_{0}$ and the wavelet high-frequency domain $H_L$ to obtain the enhanced output $I_E$. At the same time, we follow previous work \cite{panagiotou2024denoising,li2023fastllve} and simplify a denoising module to optimize the enhancement result, achieving a visually friendly enhancement. This can be expressed as:
\begin{equation}
I_E=denoise(IDWT(\hat{x}_{0},H_L)).
\tag{11}
\end{equation}

\begin{table*}[t]
\caption{Quantitative evaluation of different unsupervised learning methods on benchmark datasets. The best, second and third performance are marked in {\color[HTML]{FF0000}{red}}, {\color[HTML]{00009B}{blue}} and {\color[HTML]{548235}{green}}, respectively. }
\renewcommand\arraystretch{1.5}
\scalebox{0.86}{
\begin{tabular}{c|l|c|cccc|cccc|cc}
\hline
                                    &                          &                             & \multicolumn{4}{c|}{LOL}                                                                                                    & \multicolumn{4}{c|}{SICE}                                                                                                    & \multicolumn{2}{c}{Unpaired-set}                               \\ \cline{4-13} 
\multirow{-2}{*}{Type}              & \multirow{-2}{*}{Models} & \multirow{-2}{*}{Reference} & PSNR $\uparrow$                         & SSIM $\uparrow$                        & LPIPS $\downarrow$                        & FID $\downarrow$                           & PSNR $\uparrow$                         & SSIM $\uparrow$                        & LPIPS $\downarrow$                        & FID $\downarrow$                            & MUSIQ $\uparrow$                        & LOE $\downarrow$                           \\ \hline
                                   & Enlightengan \cite{jiang2021enlightengan}             & TIP'21                      & 17.873                        & 0.653                        & 0.393                        & 102.869                       & {\color[HTML]{548235} 18.225} & 0.769                        & 0.394                        & 110.772                        & 59.736                        & 452.524                        \\
                                    & CLIP-Lit \cite{liang2023iterative}                 & ICCV'23                     & 12.714                        & 0.481                        & 0.394                        & 120.800                       & 11.968                        & 0.632                        & 0.405                        & 118.071                        & {\color[HTML]{4472C4} 62.953} & 230.279                        \\
                                    & NeRco \cite{yang2023implicit}                   & ICCV'23                     & {\color[HTML]{FF0000} 22.946} & 0.773                        & {\color[HTML]{548235} 0.327} & 88.085                        & 17.977                        & 0.756                        & 0.428                        & 126.716                        & {\color[HTML]{548235} 62.851} & 203.493                        \\
                                   & PairLIE \cite{fu2023learning}                  & CVPR'23                     & 19.735                        & {\color[HTML]{548235} 0.776} & 0.357                        & 98.139                        & {\color[HTML]{FF0000} 19.985} & {\color[HTML]{2F75B5} 0.777} & {\color[HTML]{4472C4} 0.370} & 119.541                        & 62.229                        & 198.143                        \\
\multirow{-5}{*}{Unpaired Training} & LightenDiffusion \cite{jiang2024lightendiffusion}         & ECCV'24                     & {\color[HTML]{2F75B5} 21.099} & {\color[HTML]{FF0000} 0.821} & {\color[HTML]{4472C4} 0.310} & 93.784                        & {\color[HTML]{4472C4} 19.086} & {\color[HTML]{548235} 0.776} & {\color[HTML]{548235} 0.375} & 103.232                        & 61.233                        & {\color[HTML]{548235} 178.594} \\ \hline
                                    & Zero\_dce \cite{guo2020zero}               & CVPR'20                     & 15.053                        & 0.542                        & 0.381                        & 95.571                        & 16.076                        & 0.732                        & 0.411                        & 111.878                        & 57.439                        & 217.668                        \\
                                     & Zero\_dce++ \cite{li2021learning}           & TPAMI'21                    & 14.682                        & 0.522                        & 0.407                        & 87.552                        & 16.081                        & 0.735                        & 0.421                        & 120.410                        & 58.559                        & 324.096                        \\
                                     & RUAS \cite{liu2021retinex}                    & CVPR'21                     & 16.504                        & 0.488                        & 0.395                        & 116.757                       & 14.876                        & 0.703                        & 0.442                        & 128.073                        & 52.810                        & 402.819                        \\
                                    & SCI \cite{ma2022toward}                     & CVPR'22                     & 14.651                        & 0.502                        & 0.372                        & {\color[HTML]{4472C4} 84.476} & 14.770                        & 0.698                        & 0.415                        & {\color[HTML]{FF0000} 94.562}  & 58.667                        & 214.719                        \\
                                    & GDP \cite{fei2023generative}                     & CVPR'23                     & 15.896                        & 0.542                        & 0.431                        & 112.363                       & 14.878                        & 0.527                        & 0.509                        & 162.205                        & 56.217                        & {\color[HTML]{4472C4} 161.475} \\
                                    & FourierDiff \cite{lv2024fourier}             & CVPR'24                     & 18.673                        & 0.602                        & 0.362                        & {\color[HTML]{548235} 86.495} & 16.871                        & 0.768                        & 0.387                        & {\color[HTML]{4472C4} 98.583}  & 57.117                        & 200.472                        \\
\multirow{-7}{*}{Zero-Shot}         & Ours                     &   -                          & {\color[HTML]{548235} 20.922} & {\color[HTML]{4472C4} 0.811} & {\color[HTML]{FF0000} 0.281} & {\color[HTML]{FF0000} 63.601} & {\color[HTML]{548235} 18.335} & {\color[HTML]{FF0000} 0.779} & {\color[HTML]{FF0000} 0.367} & {\color[HTML]{548235} 102.277} & {\color[HTML]{FF0000} 63.241} & {\color[HTML]{FF0000} 158.470} \\ \hline
\end{tabular}
}
\label{vib}
\end{table*}

\begin{figure*}[ht!]\centering
     \includegraphics[height=0.27\textwidth,width=1\textwidth]{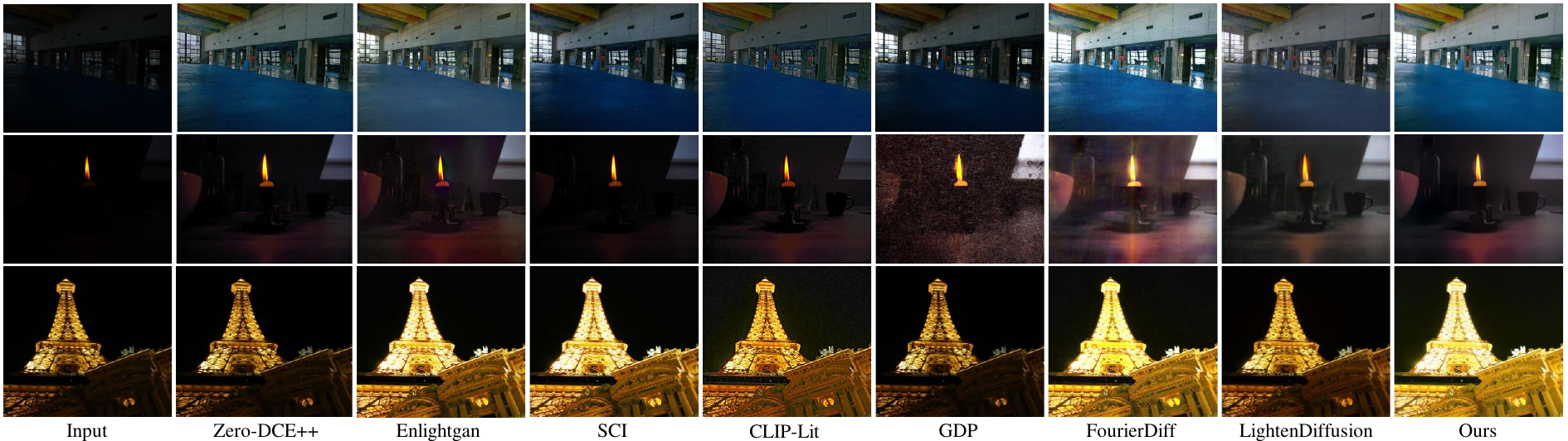}	    
     \caption{Visual comparison of the results of different methods of enhancement, best viewed by zooming in}
    \label{vis}
    % \vspace{-5mm}
\end{figure*}

\subsection{Network Optimisation}
To ensure the visual perception and luminance distribution of the sampling results at each step, we embed multimodal text in the diffusion process and supervise the enhancement process through the dominance of different modalities. Specifically, we introduce the frozen pre-trained CLIP model \cite{radford2021learning} and extract its feature vectors by feeding the preset positive prompts $T_p$ and negative prompts $T_n$ (as shown in Fig. \ref{frame}) to the text encoder $\Phi_{text}$. And the feature extraction is performed by the image encoder $\Phi_{image}$ on the sampling results $\hat{x}_t$ of each step. Subsequently, we measure the difference between image vectors and text vectors by calculating the similarity loss between them in the CLIP latent space to facilitate the alignment of the image feature space through the semantic bootstrapping ability, formulated as:
\begin{equation}
\mathcal{L}_{TG}=\sum_{t\in[0,T]}\frac{e^{\cos(\Phi_{image} (\hat{x}_t ),\Phi_{text}(T_n ))} }{\sum _{j\in\{T_p,T_n\}} e^{\cos(\Phi_{image} (\hat{x}_t),\Phi_{text}(T_j))}}.
\tag{12}
\end{equation}

Since $\vartheta$ is a learnable parameter, to optimise it during the enhancement process, we introduce the non-reference luminance control constraint $\mathcal{L}_{bri}$ at each step, which is defined as:
\begin{equation}
\mathcal{L}_{bri}=\frac{1}{N}\sum_{n=1}^{N}\parallel I_t^n-E\parallel,
\tag{13}
\end{equation}
where $N$ is the total number of non-overlapping local regions of size $16 \times 16$. $I^n_t$ denotes the average intensity value of local region $n$ in the updated sampling result $\hat{x}_t^1$. $E$ denotes the brightness level.

\section{Experiments}
\subsection{Experimental Settings}
\subsubsection{Datasets and Metrics}
We select test images from two paired datasets, LOL \cite{wei2018deep,yang2020fidelity} and SICE \cite{cai2018learning}, to evaluate the performance of the different methods. Additionally, we gather low-light images from the LIME \cite{guo2016lime}, DICM \cite{lee2013contrast}, and MEF \cite{ma2015perceptual} datasets. This set is simply called the “Unpaired-set”. For the paired dataset, we use two distortion metrics, PSNR and SSIM \cite{wang2004image}, and two perception metrics, LPIPS \cite{zhang2018unreasonable} and FID \cite{heusel2017gans}, to evaluate the network. For the unpaired dataset, we choose two metrics, MUSIQ \cite{ke2021musiq} and LOE \cite{wang2013naturalness}, to evaluate them.

\subsubsection{Implementation Details}
We implemented our framework on a single NVIDIA Tesla V100 GPU using PyTorch. We use an unconditional 256×256 diffusion model pre-trained on ImageNet \cite{deng2009imagenet}. The total diffusion step size T and the alternating optimisation interval step size S are set to 1000 and 200, respectively. 

\subsubsection{Quantitative Comparisons}
We performed a quantitative comparison of all reference datasets. As shown in Table \ref{vib}, in all real datasets, we achieved the top three performances compared to all competing methods. Notably, in the zero-shot method, we achieved the best performance evaluations. This fully validates the effectiveness and generalization of the model.
\subsubsection{Qualitative Comparisons}
As shown in Fig. \ref{vis}, we observed that LightenDiffusion suffers from colour distortion problems. GDP and FourierDiff are under-enhanced and have obvious artifacts. In contrast, our method has more realistic colours and image details, and thus has a visual effect more in line with human perception.

% \begin{figure}[ht!]\centering
%      \includegraphics[height=0.14\textwidth,width=0.4\textwidth]{fre.pdf}	    
%      \caption{A visual comparison of results with and without the Frequency Domain Perception Module.}
%     \label{fre}
%     \vspace{-3mm}

% \end{figure}

\begin{table}[]\centering

\caption{Ablation experiments with different modules}
\renewcommand\arraystretch{1.3}
\label{ablation}
\begin{tabular}{cccccc}
\hline
Wavelet Priori
 & Text Guide
 & PSNR$\uparrow$   & SSIM$\uparrow$  & LPIPS$\downarrow$ & FID$\downarrow$    \\ \hline
$\times$         & $\times$    & 18.585 & 0.771 & 0.309 & 76.596 \\
$\checkmark$         & $\times$    & \textcolor{red}{20.941} & 0.807 & 0.287 & 65.159 \\
$\checkmark$         & $\checkmark$    & 20.922
 & \textcolor{red}{0.811} & \textcolor{red}{0.281} & \textcolor{red}{63.601} \\ \hline
\end{tabular}
% \vspace{-5mm}

\end{table}
\subsection{Ablation Study}
To evaluate the effectiveness of our model, we performed ablation experiments on different modules on the LOL test set. Table \ref{ablation} shows that Wavelet Priori effectively steers the sampling process and constrains the generation of redundant image details. Text guide further enhances image structure and visual perception.

\section{Conclusion}
This letter proposes a novel zero-shot low-light image enhancement method , which compensates for the lack of illumination information and structural information in the zero-shot diffusion sampling process by effectively combining the wavelet domain and Fourier frequency domain to construct rich a priori information. In addition, we embed multimodal text to further facilitate image enhancement and make the backsampling process more stable. A large number of experiments verify the effectiveness and robustness of our method.

% use section* for acknowledgment

% Can use something like this to put references on a page
% by themselves when using endfloat and the captionsoff option.
\ifCLASSOPTIONcaptionsoff
  \newpage
\fi

\vfill
\pagebreak

\bibliographystyle{plain}
\bibliography{template}

% that's all folks
\end{document}